\documentclass{article}
\usepackage{graphicx} 
\usepackage{algorithm}
\usepackage{algpseudocode}
\usepackage{amsmath}
\usepackage{float}

\usepackage{listings}
\usepackage{xcolor}

\usepackage{booktabs}
\usepackage{longtable}

\usepackage{hyperref}
\usepackage[symbol]{footmisc}

\lstdefinelanguage{json}{
    basicstyle=\ttfamily\small,
    backgroundcolor=\color{lightgray!20},
    stringstyle=\color{purple}, 
    numberstyle=\color{blue},  
    keywordstyle=\color{red}\bfseries, 
    morestring=[b]",
    literate=
     *{0}{{{\color{blue}0}}}{1}%
      {1}{{{\color{blue}1}}}{1}%
      {2}{{{\color{blue}2}}}{1}%
      {3}{{{\color{blue}3}}}{1}%
      {4}{{{\color{blue}4}}}{1}%
      {5}{{{\color{blue}5}}}{1}%
      {6}{{{\color{blue}6}}}{1}%
      {7}{{{\color{blue}7}}}{1}%
      {8}{{{\color{blue}8}}}{1}%
      {9}{{{\color{blue}9}}}{1}%
      {:}{{{\color{black}:}}}{1}
      {,}{{{\color{black},}}}{1}
      {true}{{{\color{red}true}}}{4}%
      {false}{{{\color{red}false}}}{5}%
      {null}{{{\color{red}null}}}{4}%
}

\title{ShadowGenes: Leveraging Recurring Patterns within Computational Graphs for Model Genealogy}
\author{
    \makebox[0.5\linewidth][c]{Kasimir Schulz} \hfill \makebox[0.5\linewidth][c]{Kieran Evans} \\ 
    \makebox[0.5\linewidth][c]{\texttt{kschulz@hiddenlayer.com}} \hfill \makebox[0.5\linewidth][c]{\texttt{kevans@hiddenlayer.com}} 
}
\date{}

\begin{document}

\maketitle

\begin{abstract}
Machine learning model genealogy enables practitioners to determine which architectural family a neural network belongs to. In this paper, we introduce ShadowGenes, a novel, signature-based method for identifying a given model’s architecture, type, and family. Our method involves building a computational graph of the model that is agnostic of its serialization format, then analyzing its internal operations to identify unique patterns, and finally building and refining signatures based on these. We highlight important workings of the underlying engine and demonstrate the technique used to construct a signature and scan a given model. This approach to model genealogy can be applied to model files without the need for additional external information. We test ShadowGenes on a labeled dataset of over 1,400 models and achieve a mean true positive rate of 97.49\% and a precision score of 99.51\%; which validates the technique as a practical method for model genealogy. This enables practitioners to understand the use cases of a given model, the internal computational process, and identify possible security risks, such as the potential for model backdooring.
\end{abstract}

\section{Introduction}

As evidenced by the fact that the number of repositories in the HuggingFace model zoo recently surpassed one million \footnote{\url{https://huggingface.co/models}}, it is clear that pre-trained machine learning models are being developed and made available for public use at a rapid rate. The free exchange of pre-trained models comes with a myriad of benefits. However, there are significant risks introduced by this practice that are in need of mitigation. One such risk, and the focus of this paper, is that practitioners may download and run pre-trained models that were not described accurately by the model’s publisher, either by accident or due to malicious intent. Those who wish to leverage a pre-trained neural network downloaded from an open model repository currently have limited ways in which they can verify the network architecture (e.g., convolutional neural network, transformer, LSTM) and modality (e.g., image, text, audio, video).

There are several terms used within this whitepaper that have similar meanings but have distinct definitions, as set out here. Within the scope of this paper, when we reference a model’s architecture, we are referring to the inner operations of the model in a more general sense; an example would be a convolutional neural network (CNN) or a transformer. When we refer to a model’s family, this is a more specific term; an example would be the ResNet family, which leverages a CNN architecture. When we specify model type, it pertains to the task of the model; an example type would be an image recognition model - the family of which could be ResNet (CNN architecture) or VisionTransformer (transformer architecture). When we refer to a derivative of a model, this is a model family that builds upon the architecture of another; an example would be RoBERTa being a derivative of BERT.

It is critical for consumers of model repositories to be able to quickly, easily, and reliably determine whether or not they are downloading a neural network with the expected architecture and data modalities. At the time of writing, there are no regulations in place for this; however, there are several recommendations within NIST's AI Risk Management Framework relating to the trustworthiness of AI systems that pertain to their validity, reliability, security and transparency \footnote{\url{https://airc.nist.gov/AI_RMF_Knowledge_Base/AI_RMF/Foundational_Information/3-sec-characteristics}}. Model genealogy allows organizations to align with this by enforcing internal policy decisions around models. 

In this paper, we introduce ShadowGenes, a novel method for identifying model architectures, families, and specific model types. This method can help validate a model by confirming that its internal mechanism corresponds with the desired task and ensure the reliability of the model by highlighting its heritage and architecture. This approach also enables practitioners to understand potential security risks associated with a model. Our previous work has shown CNN models are substantially more vulnerable to model backdooring than VIT models \cite{shadowlogic}. Therefore, automatically verifying a model’s family and architecture can substantially reduce the risk of unknowingly deploying sensitive models into security and safety critical AI systems. This technique can assist with transparency in that it identifies the model family and type, and can therefore aid in understanding the model’s operational steps taken all the way from the model’s input to its output.

\section{Related Work}

DNN Genealogy enables practitioners to visualize multiple aspects of supported deep neural networks, including their architecture and the evolutionary relationships between them \cite{8732351}. Within this visualization tool, links between models and their derivatives are made and described, with the nodes of the graph being a model family (such as AlexNet) and the edges being links between parents and children, which includes the strength of the link and an explanation of it (for example AlexNet to VGG).

PhyoLM is a model genealogy tool for Large Language Models (LLMs). In addition to describing relationships between LLMs, PhyoLM predicts a model’s performance based on its relationships with other LLMs \cite{yax2024phylolminferringphylogeny}.

ShadowGenes differs from these tools in that it: 

\begin{enumerate}
    \item Builds a format-agnostic computational graph of a model;
    \item Groups its nodes and edges into semantically meaningful blocks, and,
    \item Runs a pattern-matching algorithm to compare repeated patterns within the graph (which we refer to as subgraphs) to a pre-constructed database of signatures that are necessary and sufficient for assigning a model to a particular family. 
\end{enumerate}

This approach achieves a high degree of success in determining a model’s intended tasks (including the identification of more than one modality), the family it belongs to, its derivatives, and relationships with other models. To expand on this by way of example, RoBERTa and DistilBERT are model families that both derive from the natural language processing (NLP) model BERT, so they are considered to be related to each other through the use of this architecture. Additionally, ShadowGenes can be applied to models without any requirement for external information or metadata.

\section{Methodology}

\subsection{Extracting a Model Format Agnostic Graph}

ShadowGenes works for models that have been serialized using any format that includes a computational graph representation of the model, such as ONNX, TensorFlow, CoreML, OpenVino, or Caffe. ShadowGenes' signature scanning technique only requires limited information from the computational graph and is completely independent of the underlying serialized model format or model framework. Therefore, the first step in the process is to create a format-agnostic computational graph that contains the essential information required for the subsequent steps. Three classes are used to represent the graph: Graph, Edge, and Node, which are shown in Appendix A.

While each model format stores its computational graph differently, when interacting with the serialized model file, development frameworks, such as ONNX and TensorFlow, typically expose an interface to access and extract the nodes and edges of the computational graph. This means that the agnostic graph can be constructed using the algorithm outlined below. 

\begin{algorithm}[H]
    \caption{Agnostic Graph Creation Algorithm}
    \begin{algorithmic}
        \State Function \textbf{ConstructAgnosticGraph}(\textit{Model}):

        \State \quad \quad Graph := new Graph()
        \State \quad \quad NodeLookup := new $<$String, Node$>$()

        \State

        \State \quad \quad For operation in Model.graph.nodes: // Loop through the nodes in the graph
        \State \quad \quad \quad \quad Node := new Node(operation.name, operation.operation\_type) // Create a new agnostic node with the operation name, and type, and a unique id
        \State \quad \quad \quad \quad Graph.AddNode(Node)
        \State \quad \quad \quad \quad NodeLookup[Node.name] = Node // Add the node to the dictionary for easy lookup later on

        \State

        \State \quad \quad For operation in Model.graph.nodes: // Loop through all of the nodes to assign edges
        \State \quad \quad \quad \quad Node := NodeLookup[operation.name]
        \State \quad \quad \quad \quad For input in operation.inputs: // Iterate over the inputs to the node
        \State \quad \quad \quad \quad \quad \quad InputNode := NodeLookup[input.name]
        \State \quad \quad \quad \quad \quad \quad Edge := new Edge(InputNode, Node) // Create an edge from the input node to the current node
        \State \quad \quad \quad \quad \quad \quad Graph.AddEdge(Edge)
        \State \quad \quad \quad \quad For output in operation.outputs: // Iterate through the outputs of the node
        \State \quad \quad \quad \quad \quad \quad OutputNode := NodeLookup[output.name]
        \State \quad \quad \quad \quad \quad \quad Edge := new Edge(Node, OutputNode) //  Create an edge from the current node to the output node
        \State \quad \quad \quad \quad \quad \quad Graph.AddEdge(Edge)

        \State

        \State \quad \quad Return Graph
    \end{algorithmic}
\end{algorithm}

The pseudocode displayed in the above algorithm assumes that the names of the operations are unique because most computational graphs require unique names due to single static assignment \footnote{\url{https://github.com/onnx/onnx/blob/main/docs/IR.md}}; if that is not the case for a specific file format, the node lookup can be adjusted. In some formats, such as ONNX, some operations are stored under Graph.Input and Graph.Output rather than having all of the operations in the same place. If this is the case, the algorithm can be adjusted as needed to gather all of the operations required to construct the graph.

\subsection{Extracting Repeating Subgraphs}

Once the agnostic graph is created, repeated subgraphs are extracted so signatures can be created for them, either programmatically or visually. Subgraphs are based on blocks of nodes rather than nodes themselves. Therefore, the first step of the process of extracting subgraphs is to extract the blocks from the graph.

\subsubsection{Defining Linear Blocks in a Computational Graph}

A block is defined as a sequence of connected nodes where the flow of execution is linear, meaning it does not converge from multiple inputs or diverge into multiple outputs. When constructing blocks from computational graphs, there are three types of execution flow that need to be considered: linear, diverging, and converging execution flows.

A linear execution flow constitutes a block because it has no branching, and all its nodes can be grouped into a block. Diverging and converging execution flows both represent either the start or the end of a block. A converging flow occurs when a node has more than one input, while a diverging flow occurs when a node has more than one output. These flows can be seen below in Figure 1.

\begin{figure}[H]
    \centering
    \includegraphics[width=0.315\textwidth]{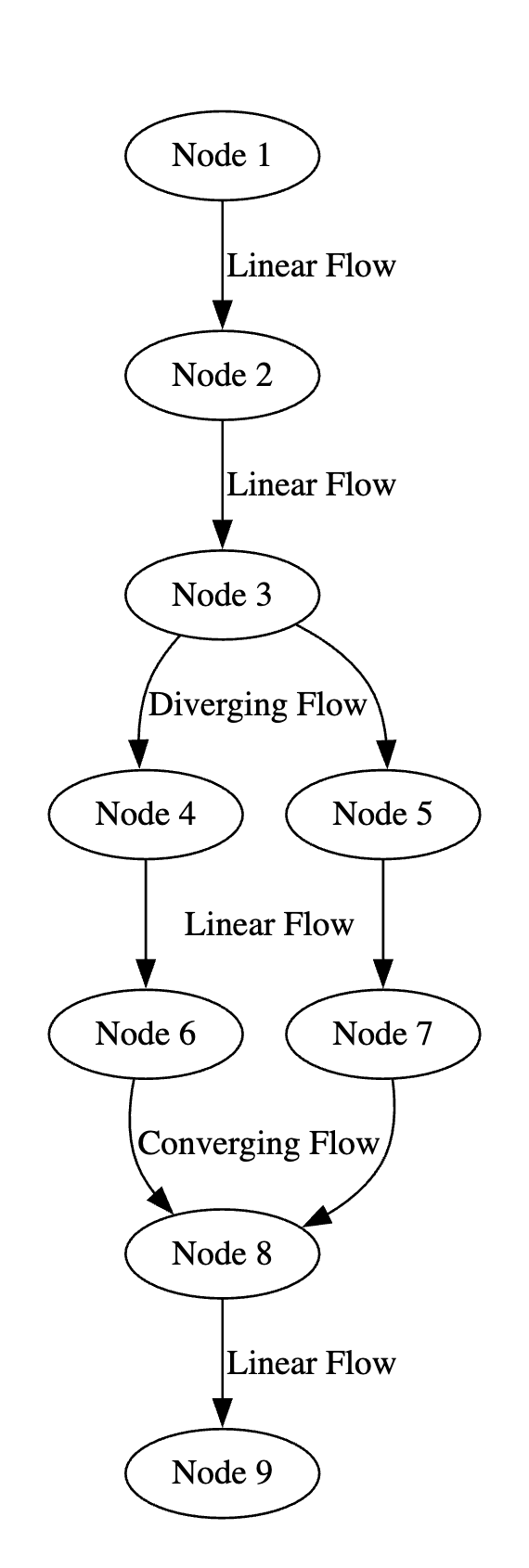}
    \caption{A figure showing the different types of execution flows.}
    \label{fig:example}
\end{figure}

In the above figure, nodes one through three are one block, as they have a linear flow. Node three is the end of this block, as it has a diverging flow. Nodes four and six comprise another block, as do nodes five and seven. Node eight receives a converging flow from nodes six and seven and has a linear flow to node nine, meaning nodes eight and nine constitute the fourth block in this subgraph.

\begin{figure}[H] 
    \centering
    \includegraphics[width=0.315\textwidth]{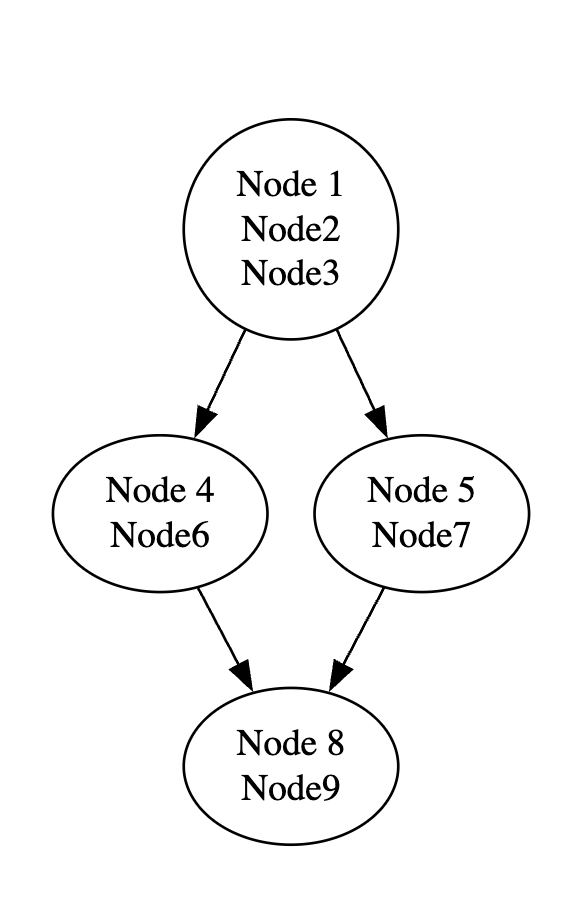}
    \caption{The different types of execution flow when blocking is applied.}
    \label{fig:example}
\end{figure}

Breaking the computational graph into blocks describing linear flow means that rather than iterating over individual nodes to scan the computational graph for a signature, blocks are used, which significantly improves the efficiency of the process because the number of objects to be iterated through is reduced due to the grouping. Using blocks also creates unique objects that are easier to scan for when creating signatures.

\subsubsection{Extracting Blocks from the Graph}

To extract the blocks and the edges between them, the ExtractBlocks algorithm is used. The algorithm is run over the format-agnostic graph created in section 3.1, ensuring it works regardless of the model file format. The ExtractBlocks algorithm works by iterating over the input nodes for the computational graph, because all input nodes will be the start of a block and will eventually go to an output node, iterating through the majority of the graph. While iterating through the nodes, the algorithm utilizes a depth-first search to traverse the nodes and create unique blocks. 

After iterating through all input nodes and their descendants to create blocks, the algorithm iterates through all remaining nodes, and if a node has not been visited previously, the depth-first search process will begin at that node. In computational graphs, a constant value or initializer will sometimes create a branch that converges with another branch that starts from an input node. The branches created by the constant value or initializer will not be found by the searches started on the input nodes. This concept is visualized in Figure 3, which shows how the first pass would find Block 1 and Block 2, but it would not find Block 3, even though its branching was considered in the creation of Block 2 because of the converging flow.

\begin{figure}[H] 
    \centering
    \includegraphics[width=0.5\textwidth]{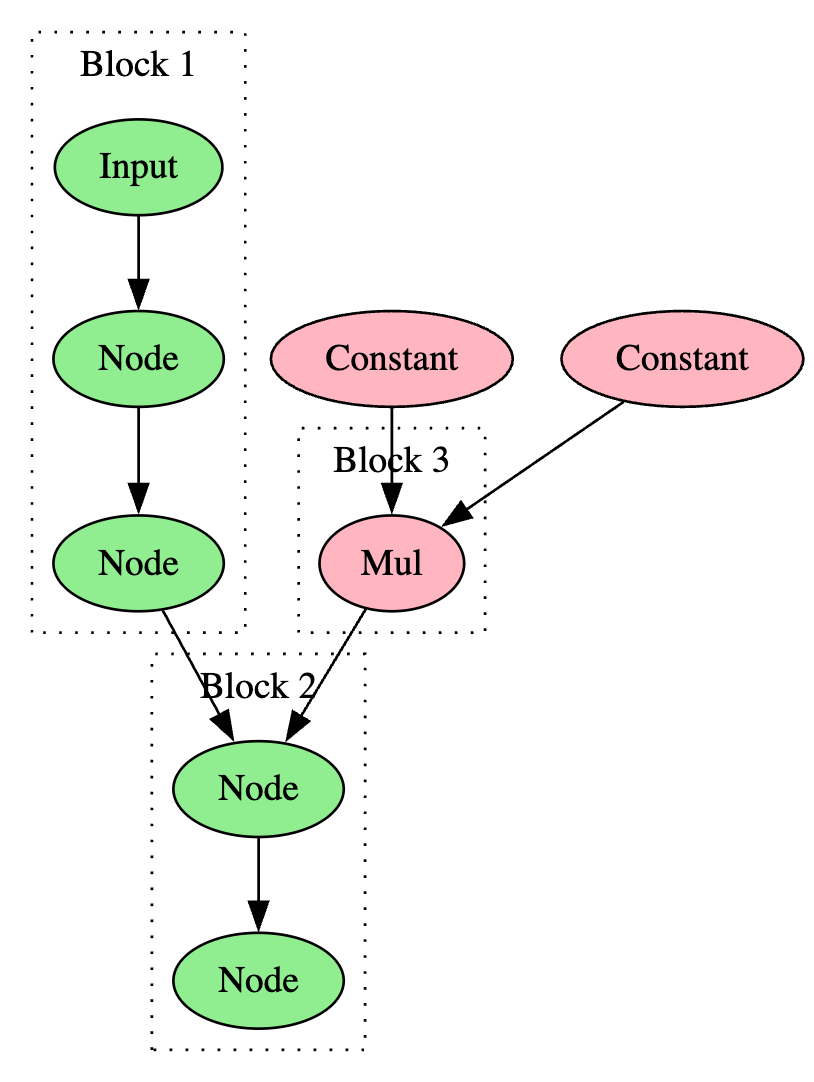}
    \caption{Blocks found by first pass.}
    \label{fig:example}
\end{figure}

In the ExtractBlocks algorithm, constant nodes are ignored as many different types of operations rely on them but they cannot be used to disrupt the execution flow. The removal of constant nodes means that there are fewer interruptions to the creation of blocks and there are fewer objects to scan through when matching signatures. In the above figure, the Mul instruction would be a block by itself while the Constant operations are ignored.

\begin{figure}[H] 
    \centering
    \includegraphics[width=0.3\textwidth]{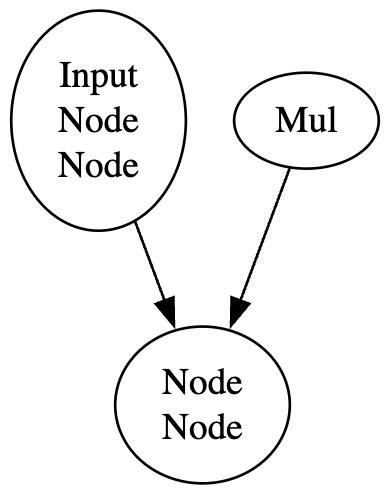}
    \caption{Blocks with Constants removed.}
    \label{fig:example}
\end{figure}

Once all blocks have been extracted from the computational graph, the ExtractBlocks algorithm loops through and identifies connections between them in order to create their respective edges. The complete ExtractBlocks algorithm is presented in Appendix B.

\subsection{Creating Signatures for Computational Graphs}

A distinguishing pattern needs to be identified to build a signature for a specific purpose, such as identifying a model family or an overarching architecture. Analyzing a visualization of a model's graphical representation can help achieve this. 

\subsubsection{The Etymology of a Signature}

Signatures are stored in JSON format so they can be quickly and easily edited and loaded into the ShadowGenes engine. Each JSON object uses the detection name as the key and must have three fields: blocks, edges, and min\_repeats.  

\begin{lstlisting}[language=json, caption={Partial Deberta Model Signature}]
"DebertaModel": {
    "blocks": [
            "id": 0,
            "ops": [
                "ConstantOfShape"
            ],
            "repeats": [1,1]
        },
        {
            "id": 1,
            "ops": [
                "Mul",
                "Equal"
            ],
            "repeats": [1,1]
        },
        ...
        {
            "id": 4,
            "ops": [
                "GatherElements",
                "*"
            ],
            "repeats": [1,1]
        }
    ],
    "edges": [
        {
            "src": 0,
            "dst": 1,
            "min_repeats": 1
        },
        ...
        {
            "src": 3,
            "dst": 4,
            "min_repeats": 1
        }
    ],
    "min_repeats": 1
}
\end{lstlisting}

Each block in the block’s array is representative of a linear execution flow block that must be present for the signature to match correctly. Each block contains an incremental integer ID, starting at zero. The operations in the block are defined under the ‘ops’ keyword and consist of either the name of an operation, a ‘?’ for single wildcard values, or a ‘*’ for multiple wildcard values. The wildcard values can be used to broaden the signatures so small changes do not affect its ability to match subgraphs. To match any singular value that needs to exist, the ‘?’ wildcard is used. To match between zero and infinity unknown values, the ‘*’ wildcard is used. The ‘repeats’ keyword is used to show the minimum and maximum number of times the operations can repeat depth-wise. For example, if a node has the ‘repeats’ value set to [1,2] and contains the Mul and Add operations, then the signature will match on either $Mul \rightarrow Add$ or $Mul \rightarrow Add \rightarrow Mul \rightarrow Add$. The complete `DebertaModel' signature is shown in Appendix D.

The edges represent how blocks are connected throughout the repeated subgraph. Each edge object contains a source and destination that correlates with one of the block IDs defined in the blocks section. Each edge also contains a ‘min\_repeats’ value that represents the minimum number of times a specific edge has to exist (width-wise). Repeated edges can occur in subgraphs, an example of which is shown in Figure 6.

Finally, the signature contains the ‘min\_repeats’ key that symbolizes the minimum number of times the subgraph needs to exist within a model's computational graph before a detection occurs.

\subsubsection{Signature Construction Through Graph Visualization}

Before running ShadowGenes on any models, high-quality signatures for model families  must be created. Visualizing the execution flow of a model’s computational graph using a tool such as Netron enables a human expert to identify its unique repeated subgraphs \footnote{\url{https://netron.app/}}. These unique elements are combined to construct a signature for a particular model, which can then be used to infer the genealogy of other models that contain the same repeated subgraph.

Figure 5 represents an identifiable subgraph observed within a ResNet50 model, a CNN used for image recognition tasks \cite{he2015deepresiduallearningimage}. We highlight three colored blocks that were leveraged during the process of building a signature for the ResNet model family.

\begin{figure}[H]
    \centering
    \begin{minipage}{0.5\textwidth}
        \centering
        \includegraphics[scale=0.5]{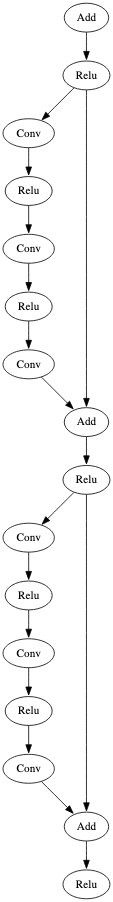}
    \end{minipage}
    \hfill
    \begin{minipage}{0.4\textwidth}
        \centering
        \includegraphics[scale=0.5]{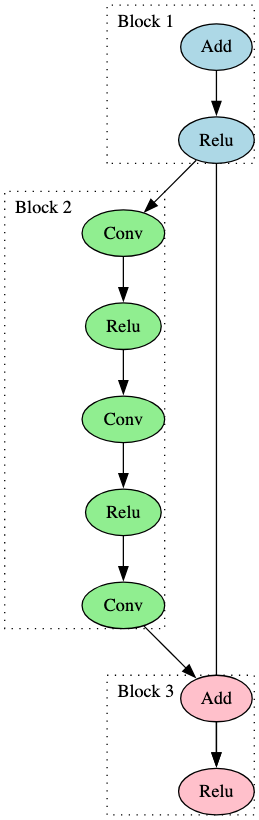}
    \end{minipage}
    \caption{An identifiable repeating pattern observed in ResNet50 models (L), and the corresponding constructed subgraph leveraged to build a signature for the ResNet model family (R).}
    \label{fig:combined layernorm}
\end{figure}

As another example, consider Inception, a CNN designed for computational efficiency, achieved by increasing the width and depth of the network \cite{szegedy2014goingdeeperconvolutions}. A recurring subgraph for a member of the Inception family is shown in Figure 6 below. In this Figure, we highlight four blocks colored blue, green, pink, and yellow, which make up the signature for the Inception model family.

\begin{figure}[H] 
    \centering
    \includegraphics[scale=0.5]{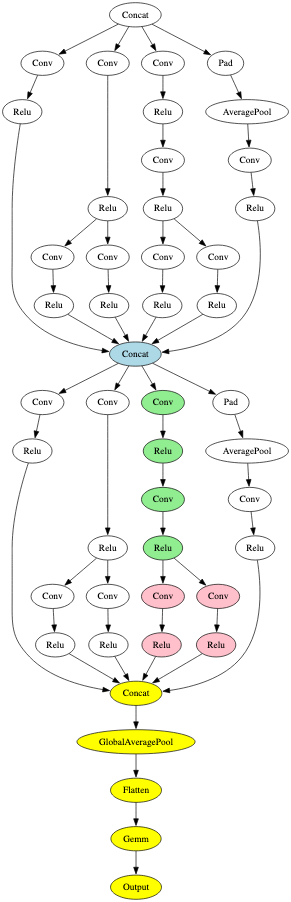}
    \caption{Depth-wise identifiable repeating pattern observed in InceptionV3 models, with the blocks used for the signature highlighted. The nodes in pink highlight the signature's repeated edge.}
    \label{fig:example}
\end{figure}

In the ResNet and Inception examples, the observed distinguishing recurring subgraphs utilized for constructing signatures repeated depth-wise through different layers. DeBERTa (Decoding-enhanced BERT with disentangled attention) is an NLP model built as an improvement on BERT and RoBERTa - subsection 5.0.2 goes into more detail on BERT derivatives - with one particular change made in the encoder layers in relation to disentanglement of the attention mechanism \cite{he2021debertadecodingenhancedbertdisentangled}. The signature for the DeBERTa model family is constructed from a recurring subgraph that repeats width-wise, as highlighted in Figure 7. 

\begin{figure}[H] 
    \centering
    \includegraphics[scale=0.4]{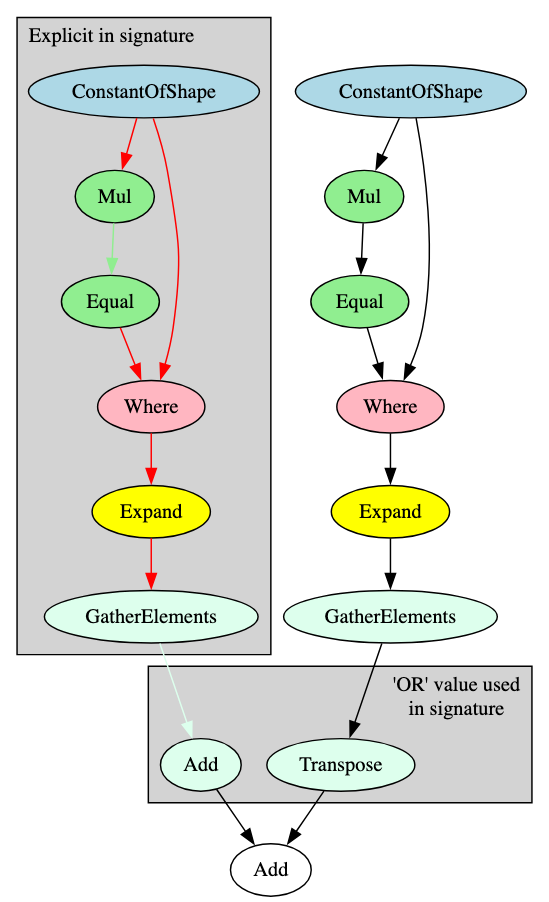}
    \caption{Width-wise identifiable recurring subgraph observed in DeBERTa, each branch is intrinsically connected to the self-attention mechanism within the layer.}
    \label{fig:example}
\end{figure}

The signature for this subgraph is comprised of four blocks. The edges highlighted in red show a connection from one block to another; the nodes and edges of the same color represent connections between nodes in the same block. In each of the width-wise repeated patterns, the `Expand' and `GatherElements' operators receive input from different attention mechanism branches within the same layer and are, therefore, one-node blocks. The `GatherElements' operators either pass to an `Add' operator or a `Transpose' operator, which is reflected in the signature with an `\textbar\textbar' (or) value. These then converge with the `Add' operator at the end, which is not part of the signature but continues the flow through the attention mechanism part of the layer. 

Once the signatures are created, they can be run using the CheckSignature Algorithm detailed in Appendix C.

\section{Experimental Results}

\subsection{Efficacy of Signatures for Detecting Model Families for Known Families}

Experimentation was carried out against models of the ONNX format, an intermediate representation that enables developers to work with different model formats in a standardized way. 

The ONNX model zoo, hosted on GitHub, contains pre-trained models sorted into ‘classes’, which correspond to the concept of model families that have been used throughout this paper \cite{onnx_repo}. In total, there are 1,483 individual verified models across 104 model families in the ONNX model zoo. The signatures that we have created cover 101 of these families. ShadowGenes was tested on the models from these families to determine how well these signatures work. For this experiment, ShadowGenes only assigned each model to a single family. 

Several classes defined within the ONNX model zoo contain multiple families. For example, ByobNet (Bring-your-own-Blocks) is used to implement multiple networks and, as a result, there are some models classified under ByobNet that were found to be of different families, including ResNet, RegNet, and GerNet. This had a negative impact on the efficacy of the signatures, in particular causing many false positives; ResNet signatures were generating false positives on ByobNet because of the ResNet models classified as ByobNet. This was remedied by leveraging the ONNX model zoo’s naming convention. Within these broader classes, such as EfficientNet and ByobNet, each model’s filename begins with a more specific descriptor of that model’s ‘sub-class’. For example, a model from the ByobNet class might have a filename beginning with ‘resnet’. We updated each model’s family label according to this naming convention, greatly improving detection accuracy.

Of the 104 model families, signatures were created for 101 of them. These signatures accurately identified 96.42\% of the 1,483 models, returned a mean true positive percentage of 97.49\%, with a precision of 99.51\% and a false positive rate of 0.47\%. A further breakdown of these results is shown in the table below.
\\

\begin{longtable}{|c|c|c|c|}
\hline
Model & True & True & False \\
Family & Positive \% & Positive Count & Positive Count \\ \hline
\endfirsthead
\endfoot
\hline
\endlastfoot
        AlexNet & 100\% & 2/2 & 0 \\
        BEIT & 100\% & 2/2 & 0 \\
        BERT & 100\% & 2/2 & 0 \\
        BigBird & 100\% & 3/3 & 0 \\
        CAIT & 100\% & 18/18 & 0 \\
        CoaT & 100\% & 12/12 & 0 \\
        ConVit & 100\% & 6/6 & 0 \\
        ConvBERT & 100\% & 3/3 & 0 \\
        ConvMixer & 100\% & 7/7 & 0 \\
        ConvNeXt & 88.46\% & 46/52 & 0 \\
        CrossVit & 100\% & 4/4 & 0 \\
        CspNet & 100\% & 33/33 & 0 \\
        DEIT3 & 100\% & 26/26 & 0 \\
        DLA & 100\% & 36/36 & 0 \\
        DPN & 100\% & 12/15 & 0 \\
        DarkNet & 100\% & 6/6 & 0 \\
        DeBERTa & 100\% & 2/2 & 0 \\
        DenseNet & 86.36\% & 19/22 & 0 \\
        DistilBERT & 100\% & 3/3 & 0 \\
        ECA-ResNet & 100\% & 12/12 & 0 \\
        EdgeNeXt & 100\% & 9/9 & 0 \\
        EfficientNet & 81.33\% & 135/166 & 4 \\
        ELECTRA & 100\% & 3/3 & 0 \\
        ESM & 100\% & 3/3 & 0 \\
        FBNet & 100\% & 3/3 & 0 \\
        FCOS & 100\% & 3/3 & 0 \\
        FeaStConv & 100\% & 3/3 & 0 \\
        Funnel & 100\% & 6/6 & 0 \\
        GPTNeoX & 100\% & 3/3 & 0 \\
        GcResNet & 100\% & 6/6 & 0 \\
        GcResNext & 100\% & 5/5 & 0 \\
        GerNet & 100\% & 9/9 & 0 \\
        GhostNet & 100\% & 2/2 & 0 \\
        GoogLeNet & 100\% & 3/3 & 0 \\
        HighResolutionNet & 100\% & 27/27 & 0 \\
        IBert & 100\% & 2/2 & 0 \\
        InceptionResnetV2 & 100\% & 6/6 & 0 \\
        InceptionV3 & 100\% & 12/12 & 0 \\
        InceptionV4 & 100\% & 3/3 & 0 \\
        LEConv & 100\% & 1/1 & 0 \\
        LevitDistilled & 100\% & 8/8 & 0 \\
        LongT5Encoder & 100\% & 2/2 & 0 \\
        Longformer & 100\% & 3/3 & 0 \\
        Luke & 100\% & 1/1 & 0 \\
        MNASNet & 53.33\% & 8/15 & 0 \\
        MPNet & 100\% & 3/3 & 0 \\
        MT5Encoder & 100\% & 2/2 & 0 \\
        MlpMixer & 75\% & 18/24 & 0 \\
        MobileNetV2 & 75\% & 9/12 & 0 \\
        MobileNetV3 & 95.74\% & 45/47 & 0 \\
        MobileVIT & 100\% & 18/18 & 0 \\
        NASNetALarge & 100\% & 3/3 & 0 \\
        Nest & 100\% & 8/8 & 0 \\
        Nezha & 100\% & 3/3 & 0 \\
        NormFreeNet & 95.65\% & 22/23 & 0 \\
        Nystromformer & 100\% & 3/3 & 0 \\
        Pegasus & 100\% & 3/3 & 0 \\
        PNASNet5Large & 100\% & 3/3 & 0 \\
        PoolingVisionTransformer & 100\% & 24/24 & 0 \\
        RegNet & 94.29\% & 99/105 & 0 \\
        RepVGG & 100\% & 24/24 & 0 \\
        Res2Net & 100\% & 18/18 & 0 \\
        ResGatedGraphConv & 100\% & 1/1 & 0 \\
        ResNet & 92.94\% & 250/269 & 3 \\
        ResNetV2 & 92.59\% & 25/27 & 0 \\
        ResNext & 100\% & 3/3 & 0 \\
        RetinaNet & 100\% & 3/3 & 0 \\
        RexNet & 100\% & 8/8 & 0 \\
        RoFormer & 100\% & 1/1 & 0 \\
        RoBERTa & 88.89\% & 8/9 & 0 \\
        RoBERTaPreLayerNorm & 100\% & 3/3 & 0 \\
        SAGEConv & 100\% & 3/3 & 0 \\
        SENet & 90.48\% & 38/42 & 0 \\
        SK-ResNet & 100\% & 4/4 & 0 \\
        SK-ResNext & 100\% & 2/2 & 0 \\
        SPNASNet & 100\% & 3/3 & 0 \\
        SeBotNet & 100\% & 2/2 & 0 \\
        SeResNet & 100\% & 2/2 & 0 \\
        SelecSls & 100\% & 9/9 & 0 \\
        Sequencer2d & 100\% & 4/4 & 0 \\
        ShuffleNetV2 & 100\% & 2/2 & 0 \\
        SqueezeBERT & 100\% & 3/3 & 0 \\
        SqueezeNet & 100\% & 6/6 & 0 \\
        SwinTransformer & 100\% & 27/27 & 0 \\
        SwinTransformerV2 & 100\% & 13/13 & 0 \\
        SwinTransformerV2Cr & 100\% & 4/4 & 0 \\
        T5Encoder & 100\% & 2/2 & 0 \\
        TAGConv & 100\% & 3/3 & 0 \\
        TNT & 100\% & 3/3 & 0 \\
        Twins & 100\% & 11/11 & 0 \\
        UMT5Encoder & 100\% & 2/2 & 0 \\
        VGG & 100\% & 48/48 & 0 \\
        Visformer & 100\% & 6/6 & 0 \\
        VisionTransformer & 84.85\% & 28/33 & 0 \\
        VisionTransformerDistilled & 100\% & 7/7 & 0 \\
        VisionTransformerRelPos & 71.43\% & 10/14 & 0 \\
        VovNet & 100\% & 4/4 & 0 \\
        Xception & 100\% & 3/3 & 0 \\
        XceptionAligned & 100\% & 15/15 & 0 \\
        Xcit & 100\% & 36/36 & 0 \\
        Xmod & 100\% & 3/3 & 0 \\
        \hline 
        \textbf{Totals} & \textbf{96.42\%} & \textbf{1430/1483} & \textbf{7} \\
\end{longtable}

Whilst the results are positive and demonstrative of a legitimate genealogy methodology, future work will be done to further improve the true positive rate, and test against a broader dataset to further validate the current signature database. That being said, the simplicity of the signature building process means incorporating new models and architectural changes would be easily achievable.

\subsection{Adjusting for Architectural Nuances within Classes}

In early testing, ResNet models were producing too many false negatives. Results from our analysis indicated that this was because of a slight architectural difference between smaller and larger models. ResNet models can be defined by the number of convolutional layers they contain. Examples include ResNet18, Resnet50, and Resnet152. A typical, standard ResNet model with greater than 50 such layers is depicted in Figure 5. Referring back to this figure to explain: Block 2 is a linear flow consisting of three Convolutional operators and two Relu operators. In those with less than 50, such as ResNet18, Blocks 1 and 3 remain the same between the models, but ResNet18 contains only two Convolutional operators and one Relu operator in Block 2. In order to identify this subgraph with either of these linear flows within a model, the ‘*’ wildcard value was used within the second node of the signature, greatly reducing the false negative rate.

\subsection{Adjusting for Different ONNX Opsets}

When building the signatures, it became apparent that those constructed for a specific model family would successfully identify a model compliant with ONNX opsets 17 and 18 but would result in false negatives for those compliant with ONNX opset 16 and below. This was particularly noticeable in models using transformer architecture. Upon further analysis, it was found that this was because the LayerNormalization ONNX operator was not introduced until opset 17, meaning the graph representations of opset 16 models had the LayerNormalization operator broken down into a distinct subgraph containing multiple blocks and operators, as shown in Figure 8.

\begin{figure}[H]
    \centering
    \begin{minipage}{0.4\textwidth}
        \centering
        \includegraphics[scale=0.4]{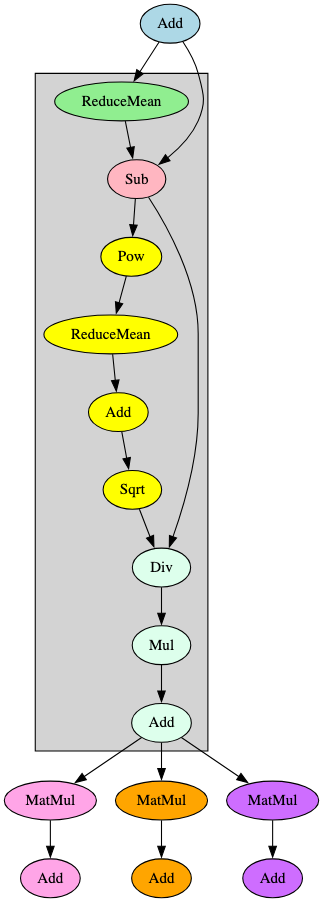}
    \end{minipage}
    \hfill
    \begin{minipage}{0.5\textwidth}
        \centering
        \includegraphics[scale=0.4]{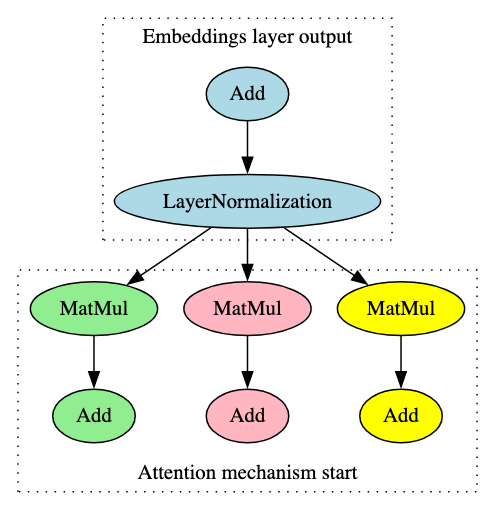}
    \end{minipage}
    \caption{LayerNormalization operation depicted in models compliant with ONNX opset 16 and below (L) and opset 17 and above (R).}
    \label{fig:combined layernorm}
\end{figure}

Both images in Figure 8 represent the flow from the embeddings layer into the first encoder layer's attention mechanism in a BERT model, the significant difference being that the LayerNormalization operator pictured on the right (opset 17) is broken down into the operators highlighted in grey on the left (opset 16). Identifying and adjusting for this difference significantly improved ShadowGenes' accuracy, with the true positive rate increasing by fifty percent in the cases of several model classes.

\subsection{Combo Signatures: Refinement through Sub-components}

Refining signatures to target a specific model family based on a single connected subgraph is not always possible, primarily because of architectural similarity between related models. To overcome this, ShadowGenes leverages what we refer to as `Combo' signatures.

Most models contain sub-components that are individually present in several model families but only observed together in one specific model. A `Combo' signature is a broader signature constructed to identify one sub-component, which, if used alone, would generate a high number of false positives. However, implementing multiple `Combo' signatures as one enables us to target specific model families or architectures based on the unique make up of their disparate sub-components.

By way of example, the signature for the VisionTransformer model family is constructed of four `Combo' signatures and will therefore trigger when this combination of its more generic sub-components are identified within the model. Two of these are highlighted in Figure 9, and the remaining two are shown in Figures 10 and 11.

\begin{figure}[H] 
    \centering
    \includegraphics[scale=0.3]{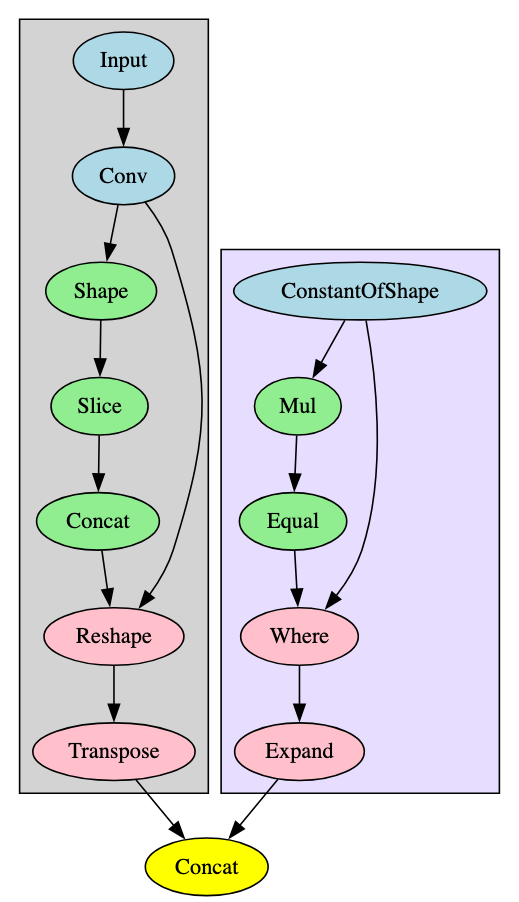}
    \caption{Two of the sub-components of the VisionTransformer model combined to construct the signature, highlighted in the grey and purple boxes.}
    \label{fig:example}
\end{figure}

The next `Combo' signature used in the VisionTransformer model identifies the activation function subsection contained within a given layer and is shown in Figure 10 below.

\begin{figure}[H] 
    \centering
    \includegraphics[scale=0.4]{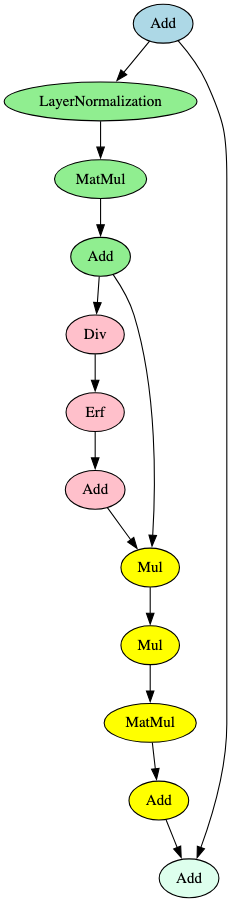}
    \caption{One of the sub-components of the VisionTransformer model combined to construct the signature.}
    \label{fig:example}
\end{figure}

The final sub-component used in the VisionTransformer model signature is one linear subgraph that leads to the output and is shown in Figure 11 below:

\begin{figure}[H] 
    \centering
    \includegraphics[scale=0.15]{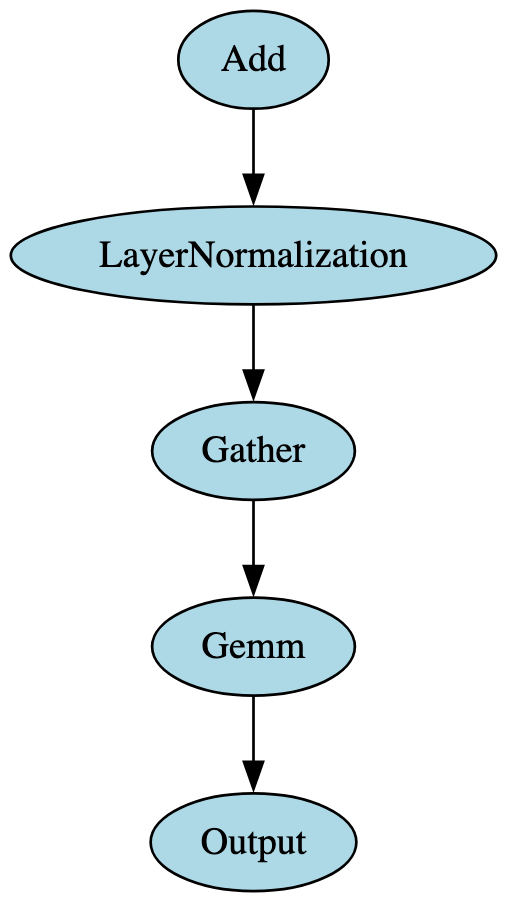}
    \caption{The final sub-component of the VisionTransformer model combined to construct the signature.}
    \label{fig:example}
\end{figure}

As stated, these subgraphs are commonly seen within multiple model families, but combining them allows us to create targeted signatures from sub-components only seen together in a particular model. A second important benefit of `Combo' signatures is that they can be reused as building blocks for many targeted signatures. For example, the activation function subgraph shown in Figure 16 is commonly seen in models that leverage the transformer architecture and can be used in conjunction with other `Combo' signatures to identify further models that use this architecture, and not just the VisionTransformer signature used here.

\subsection{Identifying Multiple Modality Components in a Single Model}

In addition to the above, further testing was conducted to determine whether or not ShadowGenes would trigger multiple signatures if applied to a multimodal model, one for each modality. When applied to a dataset of approximately 50,000 ONNX model files from HuggingFace, this was found to be the case. 

For example, an optical character recognition (OCR) model \cite{ocrmodel} triggered two signatures, demonstrating a clear delineation between its component tasks. OCR models are used to extract text from images. For example, an OCR model that performs scanned document summarization must first extract text from the scanned image (image processing) and then summarize the extracted text (natural language processing).

The ResNet signature was triggered and, as can be seen in Figure 12, is related to the model's initial input and earlier layers. 

The signature developed for the Sequencer2D models within the test set was also triggered. Sequencer leverages LSTM (Long Short Term Memory) which has also been proposed for OCR tasks because of its contextual sequencing capabilities \cite{tatsunami2023sequencerdeeplstmimage} \cite{Sabir_2017}. This part of the model is shown in Figure 12 below.

\begin{figure}[H] 
    \centering
    \includegraphics[scale=0.6]{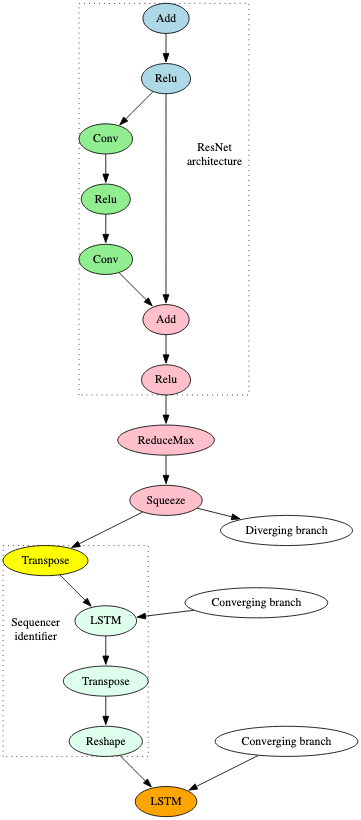}
    \caption{The ResNet layers of the OCR model pass into a Sequencer architecture, with LSTM operators observed.}
    \label{fig:example}
\end{figure}

The fact that both of these signatures triggered on the model indicates that, at the very least, the model receives an input image, performs an image recognition task, and then breaks the output down into a sequence. Knowing this is an OCR model helps us understand the component parts of the model, in that ResNet is likely used for text recognition within the image, and Sequencing / LSTM is used to contextualize the text to perform the desired task. However, even without this additional context, we could still infer that the model performs a sequencing task on an input image.

\section{Derivative Signatures}

Some model families have \textit{derivative} families that are constructed by making slight modifications to the original architecture. For example, BERT is a very popular model used for natural language processing and has several derivatives, among them BigBird, which modifies the BERT architecture to allow the model to process longer strings of textual input. Model families and their derivatives share architectural similarities, which is very useful in model genealogy as it makes it possible to determine that a model is derived from or contains components of another known model family. Knowing that a model belongs to a family that derives from another model family increases transparency, confirms the task the model was designed for, indicates how effective the model might be at its task, and can potentially flag security risks that may exist within a particular architecture. 

On the other hand, the ability to distinguish between a model and its derivatives is also a key aspect of a practical approach to model genealogy. During early rounds of testing, several false positives occurred because signatures initially designed for one model family also triggered on derivatives of that family. For example, the BERT signature was not only identifying BERT models but also its derivatives, such as BigBird and RoBERTa.

Constant signature refinement was and will continue to be necessary to account for this. Below are examples of models derived from ResNet and BERT.

\subsubsection{ResNet Derivatives}

The ResNet architecture serves as the backbone for multiple different image recognition models, including RegNet, RetinaNet, and HighResolutionNet (HRNet). 

RegNet was first proposed as a self-regulated network for image classification and is used as a regulatory module that can be appended to ResNet \cite{xu2021regnetselfregulatednetworkimage}. ShadowGenes can identify a specialized RegNet model by using a signature that contains this regulatory module. In Figure 13, the regulatory module is highlighted in grey.

\begin{figure}[H] 
    \centering
    \includegraphics[scale=0.5]{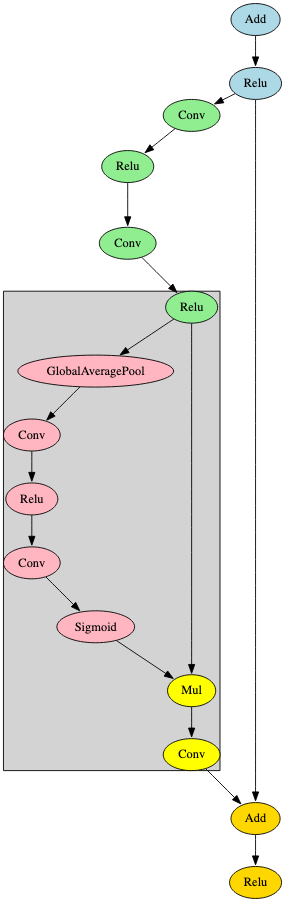}
    \caption{A key differentiator observed in RegNet models - highlighted in grey.}
    \label{fig:example}
\end{figure}

RetinaNet detects objects within images and uses ResNet as its backbone \cite{lin2018focallossdenseobject}. One of the key differentiators observed in the RetinaNet architecture is how it diverges from this backbone into several parallel sub-branches, as shown in Figure 14.

\begin{figure}[H] 
    \centering
    \includegraphics[scale=0.4]{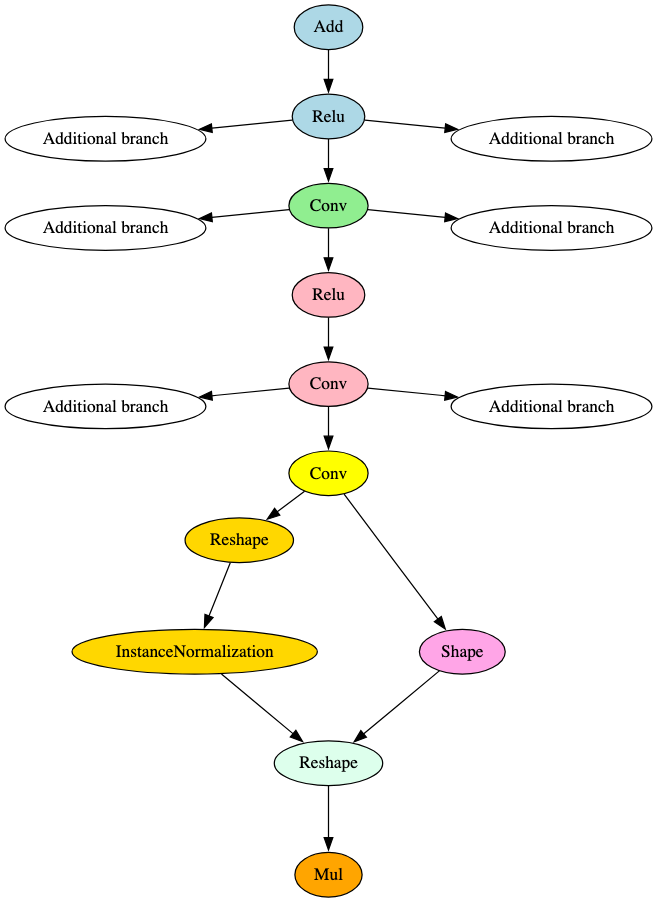}
    \caption{A key differentiator observed in RetinaNet models.}
    \label{fig:example}
\end{figure}

HighResolutionNet (HRNet) maintains high-resolution representations by connecting high-to-low-resolution convolutions in parallel \cite{sun2019highresolutionrepresentationslabelingpixels}. This is reflected in the model’s computational graph representation and can be used to build an effective signature, as shown in Figure 15.

\begin{figure}[H] 
    \centering
    \includegraphics[scale=0.45]{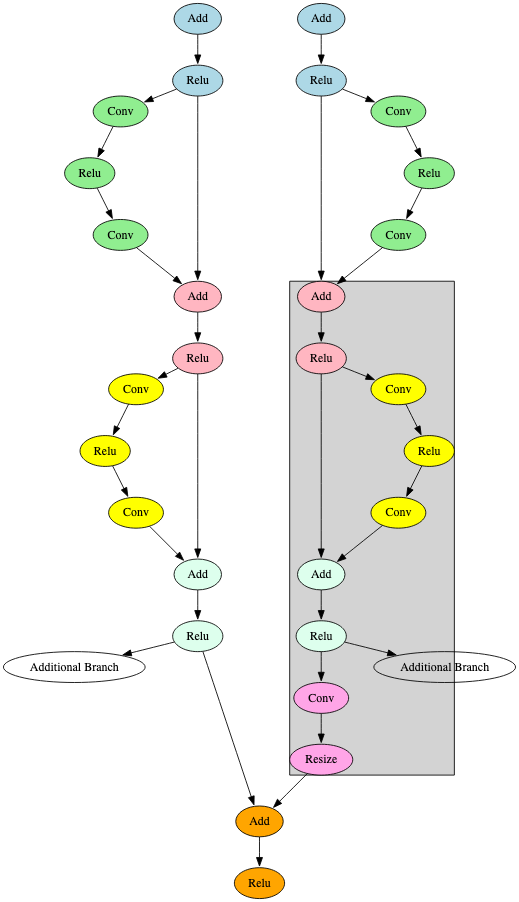}
    \caption{A key differentiator observed in HighResolutionNet models - note the parallel branches.}
    \label{fig:example}
\end{figure}

\subsubsection{BERT Derivatives}

BERT (Bidirectional Encoder Representations) is a transformer-based Language Representation Model \cite{devlin2019bertpretrainingdeepbidirectional}. It serves as the backbone for several other models that have been developed as improvements on BERT and leverage its underlying architecture. Figure 16 is a graphical representation of the attention mechanism seen within each layer of the derivatives of the BERT model described below.

\begin{figure}[H] 
    \centering
    \includegraphics[scale=0.25]{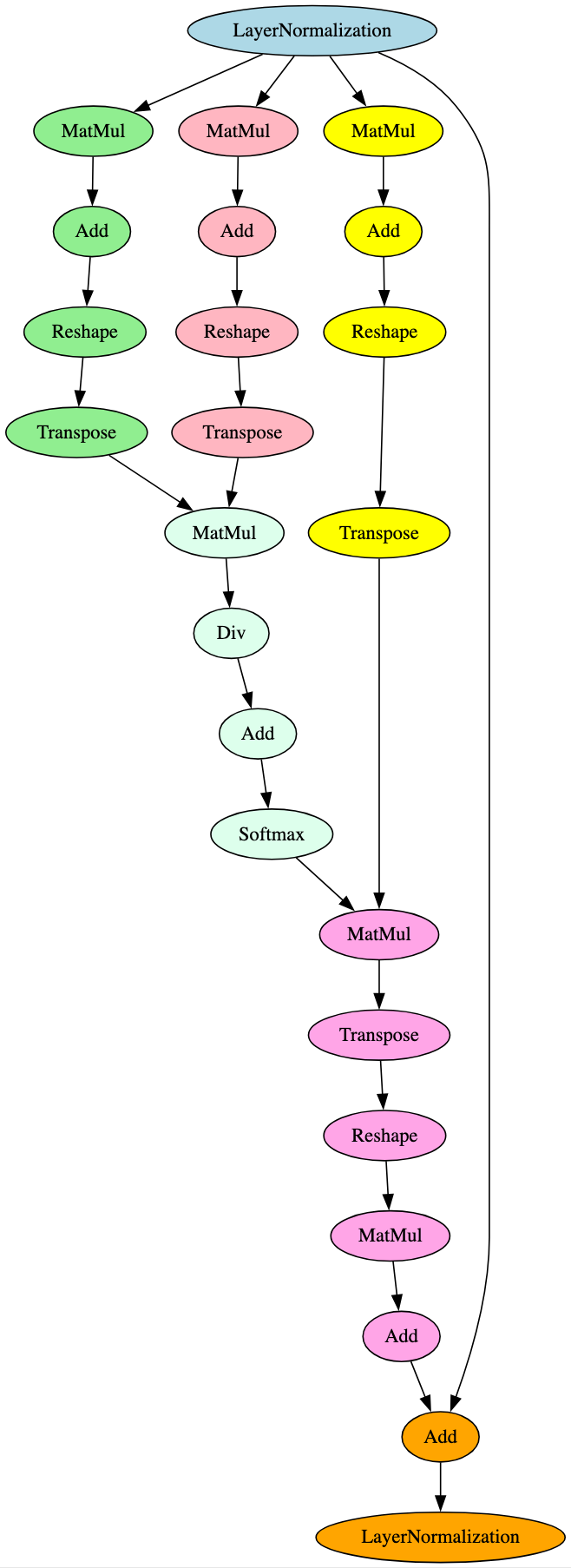}
    \caption{A graphical representation of the attention mechanism within a BERT model.}
    \label{fig:example}
\end{figure}

ELECTRA was proposed as an improvement on BERT's training method. Rather than masking input and having the model reconstruct the original tokens (MLM) during training, ELECTRA uses replaced token definition (RTD), leaving the model to identify whether the original tokens were replaced, which is more computationally efficient \cite{clark2020electrapretrainingtextencoders}. When visualizing and comparing the computational graphs for both, the observed structures are very similar. This is to be expected based on the reference paper highlighting that ELECTRA's model architecture and most hyperparameters are the same as BERT's. However, there are two key differences in the graphical representation of the models:

\begin{itemize}
    \item ELECTRA has additional operators after the LayerNormalization operation in the embeddings layer, preceding the attention mechanism in the first encoder layer; and,
    \item ELECTRA's LayerNormalization operation in the final encoder layer feeds directly to the output. With BERT, this operation diverges, with one branch passing directly into an output node, but the other passes through a pooler layer first. This aligns with the differences in training methodology.
\end{itemize}

Whilst the first item listed above is also seen in AlBERT models, the attention mechanism in the first layer is different again. Figure 17 below highlights the subgraph and differentiator leveraged for the ELECTRA signature.

\begin{figure}[H] 
    \centering
    \includegraphics[scale=0.2]{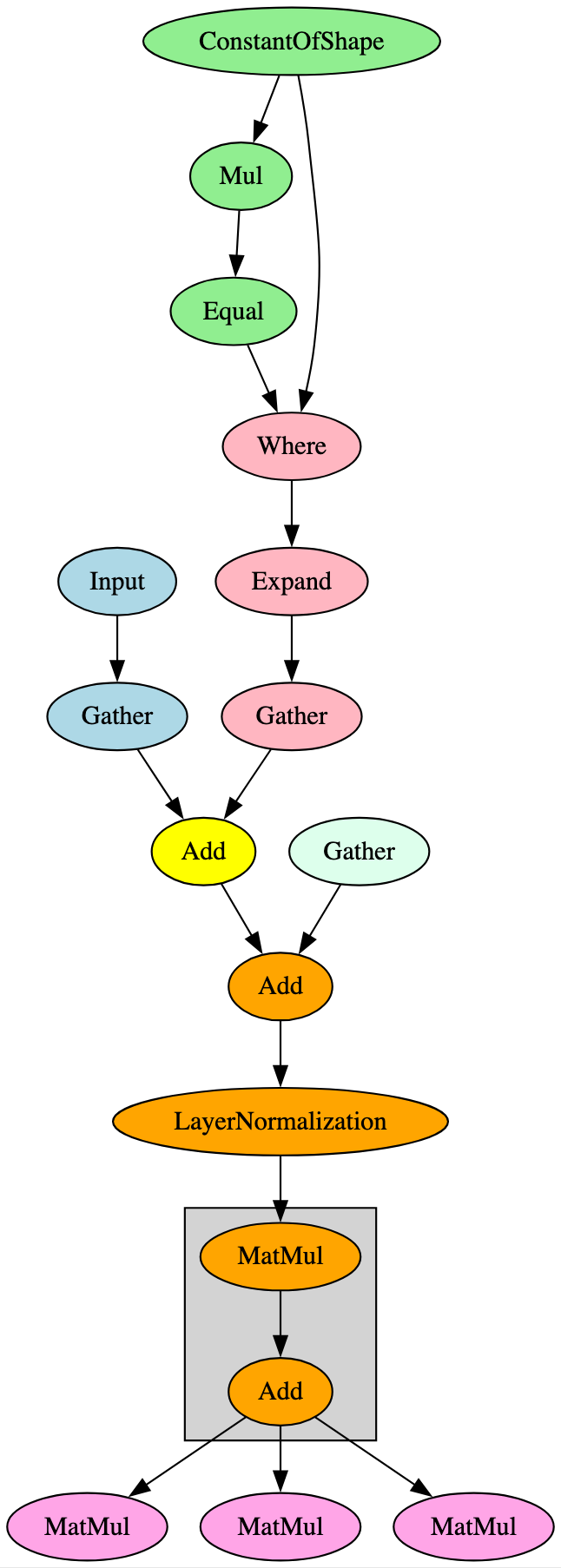}
    \caption{The differentiator following ELECTRA's LayerNormalization operation in the embeddings layer, highlighted in grey. The MatMul branches in pink are the start of the next layer's attention mechanism (see Figure 16).}
    \label{fig:example}
\end{figure}

BigBird was proposed as an improvement on BERT through the implementation of a sparse attention mechanism, allowing this variation of the model to handle longer input sequences \cite{zaheer2021bigbirdtransformerslonger}. BERT uses a full attention mechanism, making it more computationally expensive and less capable of handling longer input sequences. Despite these differences, from an architectural standpoint, the attention mechanism subsection of each layer follows the same execution flow shown in Figure 16. The architectural difference can be seen in the activation function of each layer, with those in BigBird containing additional branches, nodes, and, notably, a Tanh operator, as shown in Figure 18. Out of the models analyzed, this pattern only appeared in BigBird models.

\begin{figure}[H] 
    \centering
    \includegraphics[scale=0.25]{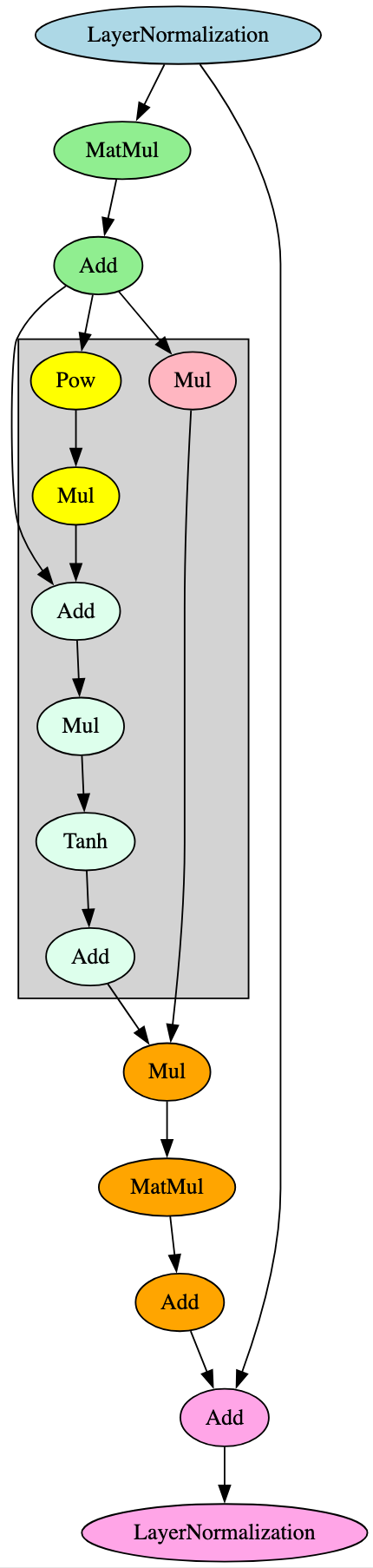}
    \caption{The differentiator in BigBird's activation function, highlighted in grey. Note: the LayerNormalization at the end feeds into the attention mechanism of the next layer, which follows the same pattern that can be seen in Figure 16.}
    \label{fig:example}
\end{figure}

Similarly to ELECTRA, RoBERTa was proposed as an improvement over BERT through alteration of the training procedure \cite{liu2019robertarobustlyoptimizedbert}. When comparing the computational graph of a RoBERTa model with BERT, they are architecturally similar, with the key differentiator observed within the embeddings layer, which has an additional branch, as shown in Figure 19. This could be explained by RoBERTa's introduction of dynamic masking, as opposed to static masking used by BERT.

\begin{figure}[H] 
    \centering
    \includegraphics[scale=0.2]{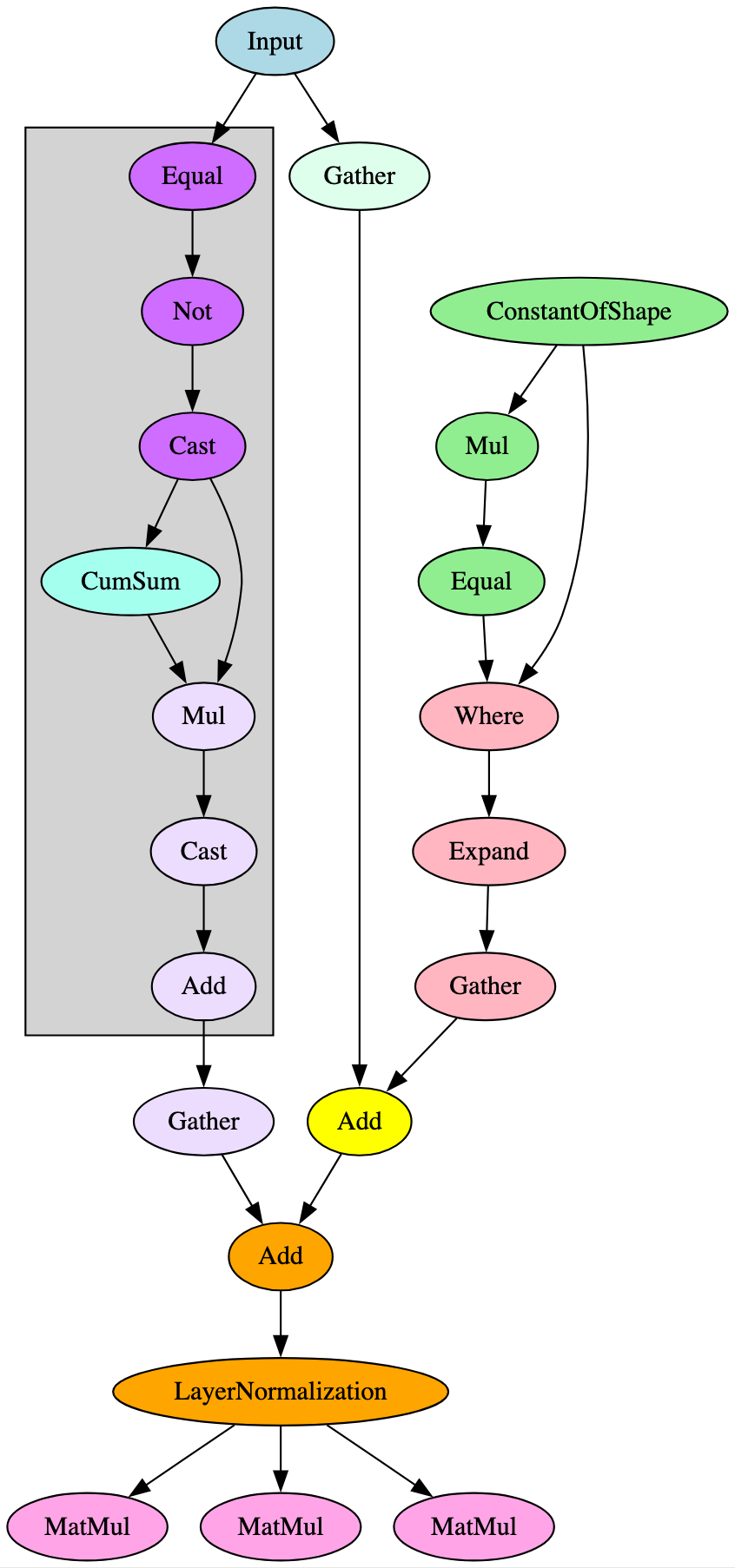}
    \caption{The differentiator in RoBERTa's embeddings layer, highlighted in grey. Note the additional branch when compared with ELECTRA's embeddings layer, shown in Figure 17. The MatMul branches in pink are the start of the next layer's attention mechanism (see Figure 16).}
    \label{fig:example}
\end{figure}

\subsubsection{Use of Broader Signatures to Identify Genealogy of Models}

For instances where ShadowGenes is not able to detect a specific model family when scanning a file, its ability to identify broader model architectures becomes important for several reasons, including:

\begin{itemize}
    \item Understanding the building blocks and backbones used within the model;
    \item Understanding the task the model was built for;
    \item Understanding any pitfalls associated with the model backbone and how there may be better alternatives for a particular use case.
\end{itemize}

The signatures outlined thus far are focused more on targeted identification of specific model types. However, with just two signatures, it was possible to identify almost all of the ResNet models within our dataset. The focus for each signature was the use of the `Conv' operator and its relationship with activation functions - one with Relu and LeakyRelu, the other with Sigmoid. Figure 20 highlights the blocks for each of these two signatures:

\begin{figure}[H]
    \begin{minipage}{0.4\textwidth}
        \raggedleft
        \includegraphics[scale=0.3]{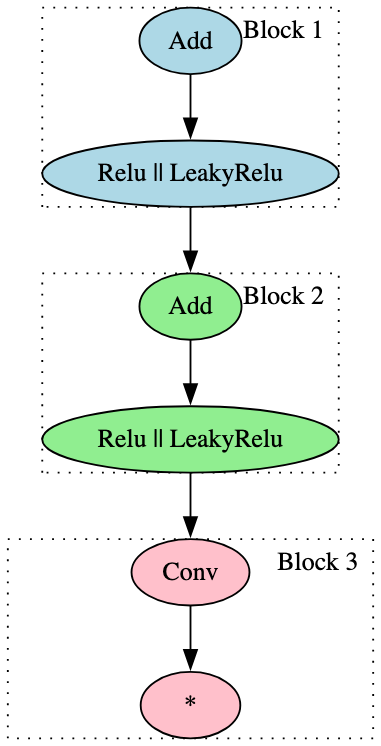}
    \end{minipage}
    \hspace{0.075\textwidth}
    \begin{minipage}{0.4\textwidth}
        \raggedright
        \includegraphics[scale=0.3]{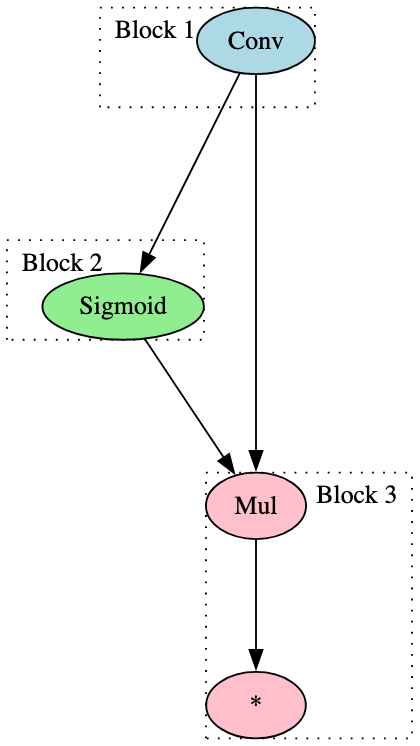}
    \end{minipage}
    \caption{Two broad signatures used to identify repeating patterns observed in ResNet model derivatives, with a focus on Relu and LeakyRelu (L), and Sigmoid (R) activation functions.}
    \label{fig:combined layernorm}
\end{figure}

In addition, the use of only two broad signatures led to the identification of many of the BERT model derivatives in the test set, each relating to the attention mask and consisting of only one block. Figure 21 highlights one of these signatures - note the only change for the other is the removal of the `Input' node, so the signature block begins with the first `Unsqueeze' node.

\begin{figure}[H] 
    \centering
    \includegraphics[scale=0.4]{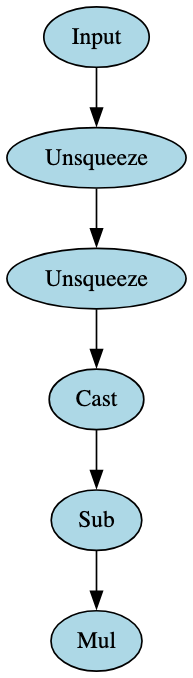}
    \caption{The full broad signature used to identify BERT model derivatives, the second signature is the same, with the `Input' node removed.}
    \label{fig:example}
\end{figure}

The broad BERT signatures detected RoBERTa models as expected. In turn, one broad signature was developed that could identify models specifically derived from RoBERTa. This is highlighted in Figure 22.

\begin{figure}[H] 
    \centering
    \includegraphics[scale=0.55]{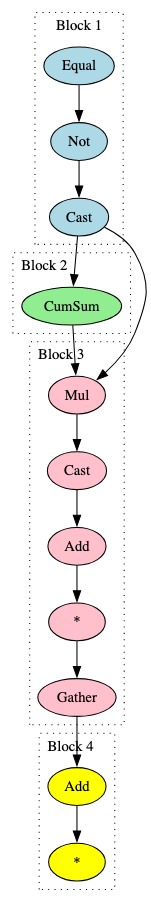}
    \caption{The broad signature used to identify RoBERTa model derivatives.}
    \label{fig:example}
\end{figure}

\section{Conclusion}

ShadowGenes allows machine learning practitioners who are considering working with an off-the-shelf model the capability to understand its genealogy and verify its use cases, reliability, and trustworthiness. In this whitepaper, we have outlined a signature-based technique for this that leverages computational graphs. We presented the results achieved when scanning a test set of models with the current signature base and how we adjusted our signatures and approach as we worked through the process. 

The results discussed here demonstrate that this methodology is an accurate and practical approach for identifying model types, families, and relationships - with or without prior relevant information. In addition, we highlight the feasibility of incorporating this approach, particularly the simplicity with which signatures can be maintained and updated. 

Not only does ShadowGenes enable practitioners to understand the ancestry, internal mechanisms, and derivatives of a given model, it is also useful in determining risks associated with model families when choosing one for deployment. This technique has also been abstracted to scanning the computational graph representation of models for branches indicative of malicious manipulation of output.

It is also important to acknowledge the limitations of this work, including the requirement to finalize signatures to account for the remaining model families and to run tests against a broader dataset. We therefore plan continual investigation and analysis to ensure the signatures remain up to date when new architectures are released or when false negatives or false positives occur.

Additionally, it is important to recognize that there are other techniques that can be applied to model genealogy. One example may include fingerprinting artifacts such as weights and biases within the layers of a model, and using the results to perform a similarity analysis between it and other models. Combining the approach outlined in this paper with other methods such as this can make model genealogy even more powerful.

We hope this paper advances understanding of model genealogy and practical solutions to it, and we invite constructive feedback and collaboration from the broader community.

\newpage

\bibliography{sources}

\clearpage
\section{Appendices}

\subsection{Appendix A - Agnostic Graph Classes}

\begin{algorithm}[H]
    \caption{Agnostic Graph Classes}
    \begin{algorithmic}
        \State \textbf{class} \textsc{Node}
        \State \quad \quad Integer: \textit{id} 
        \State \quad \quad String: \textit{name} 
        \State \quad \quad String: \textit{operation\_type}
        \State \quad \quad Node[ ]: \textit{inputs}
        \State \quad \quad Node[ ]: \textit{outputs}: 
        \State \quad \quad Edge[ ]: \textit{edges}
        
        \State
        
        \State \textbf{class} \textsc{Edge}
        \State \quad \quad Node: \textit{source}
        \State \quad \quad Node: \textit{destination}
        
        \State
        
        \State \textbf{class} \textsc{Graph}
        \State \quad \quad $<$Integer, Node$>$: \textit{nodes}
        \State \quad \quad Edge[ ]: \textit{destination}
        \State \quad \quad Node[ ]: \textit{inputs}
        \State \quad \quad Node[ ]: \textit{outputs}
    \end{algorithmic}
\end{algorithm}

\subsection{Appendix B - Extract Blocks Algorithm}

\begin{algorithm}[H]
    \caption{Extract Blocks Algorithm}
    \begin{algorithmic}
        \State \textbf{Function} ExtractBlocks(\textit{Graph}):

        \State \quad \quad Blocks := new []
        \State \quad \quad BlockEdges := new []

        \State

        \State \quad \quad InputNodes =[ 
        \State \quad \quad \quad \quad node for node in Graph.nodes
        \State \quad \quad \quad \quad if not node.operation\_type.IsInput()
        \State \quad \quad ]

        \State

        \State \quad \quad BlockSignatures := new Set()
        \State \quad \quad Visited := new Set()
        \State \quad \quad CurrentBlock := new Block()

        \State

        \State \quad \quad for InputNode in InputNodes:
        \State \quad \quad \quad \quad dfs(Graph.nodes[InputNode.id], CurrentBlock, Visited, BlockSignatures, Blocks)
        \State \quad \quad \quad \quad if not ContainsBlock(BlockSignatures, CurrentBlock):
        \State \quad \quad \quad \quad \quad \quad Blocks.Append(CurrentBlock)

        \State

        \State \quad \quad for NodeId in Graph.nodes:
         \State \quad \quad \quad \quad if not NodeId in Visited and not Graph.nodes[NodeId].IsConstant():
        \State \quad \quad \quad \quad \quad \quad if not ContainsBlock(BlockSignatures, CurrentBlock):
        \State \quad \quad \quad \quad \quad \quad \quad \quad Blocks.Append(CurrentBlock)
        \State \quad \quad \quad \quad \quad \quad \quad \quad CurrentBlock := new Block()

        \State

        \State \quad \quad for Block in Blocks:
        \State \quad \quad \quad \quad InputNode = Block.nodes.first
        \State \quad \quad \quad \quad OutputNode = Block.nodes.last

        \State

        \State \quad \quad \quad \quad for Block2 in Blocks:

        \State \quad \quad \quad \quad \quad \quad if Block2.nodes.last.id in [
        \State \quad \quad \quad \quad \quad \quad \quad \quad input.id for input in InputNode.inputs
        \State \quad \quad \quad \quad \quad \quad ]:
        \State \quad \quad \quad \quad \quad \quad \quad \quad BlockEdges.Append(new BlockEdge(Block2, Block))

        \State

        \State \quad \quad \quad \quad \quad \quad if Block2.nodes.first.id in [
        \State \quad \quad \quad \quad \quad \quad \quad \quad output.id for output in OutputNode.output
        \State \quad \quad \quad \quad \quad \quad ]:
        \State \quad \quad \quad \quad \quad \quad \quad \quad BlockEdges.Append(new BlockEdge(Block, Block2))
        
        \State

        \State \quad \quad Return Blocks, BlockEdges
        
    \end{algorithmic}
\end{algorithm}

The depth-first search visits nodes and creates new blocks based on the execution flow criteria described in section 2.2.1. If a node has already been visited the search will terminate early to avoid running unnecessary computations, if a node has not already been visited the algorithm marks it as visited. 

The ExtractBlocks algorithm checks the flow criteria by creating a new block if one of several criteria is met, as described below 

Firstly, if a node’s number of non constant inputs is not equal to one, then a new block is made. If the count is zero, this signifies the start of a new branch, and therefore the start of a new block. If the count is greater than or equal to two, this represents a converging flow on the current node, indicating a new block has started. This check is performed prior to adding a node to a block.

After adding the current node to the current block, the algorithm checks if the node has two or more outputs, signifying a diverging execution flow, and therefore the end of a block. If there is only one output node, the next check is performed in order to determine if said output node has a non constant input node count greater than or equal to two, indicating that the next node is a new block due to a converging execution flow. If either of the above conditions are met, a new block is created and the current block is added to the list of blocks. 

The algorithm then loops through all the node’s outputs and recursively calls the depth-first search again. Once the algorithm has looped through each output it creates a new block.

\begin{algorithm}[H]
    \caption{Depth-First Search Algorithm}
    \begin{algorithmic}

        \State \textbf{Function} NumberOfInputs(\textit{Node}):

        \State \quad \quad Return Length([ 
        \State \quad \quad \quad \quad input for input in Node.inputs
        \State \quad \quad \quad \quad if not input.operation\_type.IsConstant()
        \State \quad \quad ])

        \State
    
        \State \textbf{Function} dfs(\textit{Node}, \textit{CurrentBlock}, \textit{Visited}, \textit{BlockSignatures}, \textit{Blocks}):

        \State \quad \quad InputLength = Node.inputs.length
        \State \quad \quad OutputLength = Node.outputs.length

        \State
        
        \State \quad \quad if InputLength == 0 and OutputLength == 0:
        \State \quad \quad \quad \quad Visited.Add(Node.id) 
        \State \quad \quad \quad \quad Return 

        \State

        \State \quad \quad if Node.id in Visited:
        \State \quad \quad \quad \quad if not ContainsBlock(BlockSignatures, CurrentBlock):
        \State \quad \quad \quad \quad \quad \quad Blocks.Append(CurrentBlock)
        \State \quad \quad \quad \quad CurrentBlock := new Block()
        \State \quad \quad \quad \quad Return

        \State

        \State \quad \quad Visited.Add(Node.id) 

        \State

        \State \quad \quad NonConstantInputLength = NumberOfInputs(Node)

        \State

        \State \quad \quad if NonConstantInputLength $>=$ 2 or NonConstantInputLength == 0:
        \State \quad \quad \quad \quad if not ContainsBlock(BlockSignatures, CurrentBlock):
        \State \quad \quad \quad \quad \quad \quad Blocks.Append(CurrentBlock)
        \State \quad \quad \quad \quad CurrentBlock := new Block()

        \State

        \State \quad \quad CurrentBlock.Append(Node)

        \State

        \State \quad \quad if OutputLength $>=$ 2:
        \State \quad \quad \quad \quad if not ContainsBlock(BlockSignatures, CurrentBlock):
        \State \quad \quad \quad \quad \quad \quad Blocks.Append(CurrentBlock)
        \State \quad \quad \quad \quad CurrentBlock := new Block()

        \State

        \State \quad \quad if OutputLength == 1 and NumberOfInputs(Node.outputs.first) $>=$ 2:
        \State \quad \quad \quad \quad if not ContainsBlock(BlockSignatures, CurrentBlock):
        \State \quad \quad \quad \quad \quad \quad Blocks.Append(CurrentBlock)
        \State \quad \quad \quad \quad CurrentBlock := new Block()

        \State

        \State \quad \quad for neighbor in Node.outputs:
        \State \quad \quad \quad \quad dfs(neighbor, CurrentBlock, Visited, BlockSignatures, Blocks)
        \State \quad \quad \quad \quad if not ContainsBlock(BlockSignatures, CurrentBlock):
        \State \quad \quad \quad \quad \quad \quad Blocks.Append(CurrentBlock)
        \State \quad \quad \quad \quad CurrentBlock := new Block()

        \State
        
        \State \quad \quad if not ContainsBlock(BlockSignatures, CurrentBlock):
        \State \quad \quad \quad \quad Blocks.Append(CurrentBlock)
        \State \quad \quad CurrentBlock := new Block()

    \end{algorithmic}
\end{algorithm}

\begin{algorithm}[H]
    \caption{Contains Block Function}
    \begin{algorithmic}
        \State \textbf{Function} ContainsBlock(\textit{BlockSignatures}, \textit{Block}):

        \State \quad \quad if Block.nodes.length == 0:
        \State \quad \quad \quad \quad Return True

        \State

        \State \quad \quad BlockSignature = Tuple(node.id for node in Block.nodes)

        \State

        \State \quad \quad if BlockSignature in BlockSignatures:
        \State \quad \quad \quad \quad Return True

        \State

        \State \quad \quad BlockSignatures.Add(BlockSignature)
        \State \quad \quad Return False

    \end{algorithmic}
\end{algorithm}

Throughout the process outlined above, lookups need to be performed tens of thousands of times. Therefore, simple data types such as integers are used for these lookups, these are combined with sets and tuples, because using hashable types greatly improves computational efficiency.

\subsection{Appendix C - Check Signature Algorithm}

\begin{algorithm}[H]
    \caption{Check Signature Function}
    \begin{algorithmic}
        \State \textbf{Function} CheckSignature(\textit{Subgraph}, \textit{Blocks}):

        \State \quad \quad if not CheckAllBlocks(Subgraph, Blocks): // Check if the computational graph contains all blocks defined in the signature
        \State \quad \quad \quad \quad Return False

        \State

        \State \quad \quad StartNodes := GetStartNodes(Subgraph, Blocks) // Get all blocks that match the first block in the signature

        \State

        \State \quad \quad NumberOfChecks := 0

        \State

        \State \quad \quad for StartNode in StartNodes: // Loop through starting blocks and check against the signature
        \State \quad \quad \quad \quad if CheckStartNode(StartNode, Subgraph.StartNode):
        \State \quad \quad \quad \quad \quad \quad NumberOfChecks += 1
        \State \quad \quad \quad \quad \quad \quad if NumberOfChecks $>=$ Subgraph.MinRepeats: // Verify that the pattern has been found the minimum required number of times
        \State \quad \quad \quad \quad \quad \quad \quad \quad return True

        \State
        
        \State \quad \quad Return False

    \end{algorithmic}
\end{algorithm}

\begin{algorithm}[H]
    \caption{Traverse From Start Algorithm}
    \begin{algorithmic}
        \State \textbf{Function} TraverseFromStart(\textit{Subgraph}, \textit{CurrentBlock}, \textit{SubgraphNode}, \textit{VisitedEdges}, \textit{BlockEdges}):

        \State \quad \quad if SubgraphNode == Subgraph.EndNode: // Check if the last block in the subgraph has been reached
        \State \quad \quad \quad \quad Return True

        \State

        \State \quad \quad OutgoingEdges = [
        \State \quad \quad \quad \quad Edge for Edge in BlockEdges
        \State \quad \quad \quad \quad if Edge.src == CurrentBlock
        \State \quad \quad ] // Find all outgoing edges from the current block

        \State

        \State \quad \quad SubgraphEdges = [
        \State \quad \quad \quad \quad Edge for Edge in Subgraph.Edges
        \State \quad \quad \quad \quad if Edge.src == SubgraphNode.id
        \State \quad \quad ] // Find all outgoing edges in the subgraph from the current block being matched

        \State

        \State \quad \quad FinalCheck := True

        \State

        \State \quad \quad for SubgraphEdge in SubgraphEdges:
        \State \quad \quad \quad \quad Check = False
        \State \quad \quad \quad \quad SubgraphDstNode = SubgraphEdge.dst
        \State \quad \quad \quad \quad EdgeKey = (SubgraphEdge.src, SubgraphEdge.dst)
        \State \quad \quad \quad \quad if EdgeKey not in VisitedEdges:
        \State \quad \quad \quad \quad \quad \quad VisitedEdges[EdgeKey] = 0

        \State
        
        \State \quad \quad \quad \quad for Edge in OutgoingEdges:
        \State \quad \quad \quad \quad \quad \quad if SubgraphDstNode.MatchesBlock(Edge.dst)
        \State \quad \quad \quad \quad \quad \quad \quad \quad and TraverseFromStart(Subgraph, Edge.dst, SubgraphDstNode, VisitedEdges, BlockEdges):
        \State \quad \quad \quad \quad \quad \quad \quad \quad Check = True
        \State \quad \quad \quad \quad \quad \quad \quad \quad VisitedEdges[EdgeKey] += 1
        
        \State

        \State \quad \quad \quad \quad if VisitedEdges[EdgeKey] $<$ SubgraphEdge.min\_repeats - 1:
        \State \quad \quad \quad \quad \quad \quad Check = False
        
        \State

        \State \quad \quad \quad \quad FinalCheck \&= Check
        
        \State
        
        \State \quad \quad Return FinalCheck

    \end{algorithmic}
\end{algorithm}

\subsection{Appendix D - Full DeBERTa signature}

\begin{lstlisting}[language=json]
"DebertaModel": {
    "blocks": [
        {
            "id": 0,
            "ops": [
                "ConstantOfShape"
            ],
            "ignored_ops": [
                null
            ],
            "repeats": [1,1]
        },
        {
            "id": 1,
            "ops": [
                "Mul",
                "Equal"
            ],
            "ignored_ops": [
                null
            ],
            "repeats": [1,1]
        },
        {
            "id": 2,
            "ops": [
                "Where"
            ],
            "ignored_ops": [
                null
            ],
            "repeats": [1,1]
        },
        {
            "id": 3,
            "ops": [
                "Expand"
            ],
            "ignored_ops": [
                null
            ],
            "repeats": [1,1]
        },
        {
            "id": 4,
            "ops": [
                "GatherElements",
                "Add||Transpose"
            ],
            "ignored_ops": [
                null
            ],
            "repeats": [1,1]
        }
    ],
    "edges": [
        {
            "src": 0,
            "dst": 1,
            "min_repeats": 1
        },
        {
            "src": 0,
            "dst": 2,
            "min_repeats": 1
        },
        {
            "src": 1,
            "dst": 2,
            "min_repeats": 1
        },
        {
            "src": 2,
            "dst": 3,
            "min_repeats": 1
        },
        {
            "src": 3,
            "dst": 4,
            "min_repeats": 1
        }
    ],
    "min_repeats": 1
}
\end{lstlisting}

\end{document}